\documentclass[runningheads]{llncs}
\usepackage[T1]{fontenc}
\usepackage{graphicx}
\usepackage{amsmath}
\usepackage{amssymb}
\usepackage{floatrow}
\usepackage[flushleft]{threeparttable}
\usepackage{tablefootnote}
\usepackage{array}
\newcolumntype{P}[1]{>{\centering\arraybackslash}p{#1}}
\usepackage{multicol}
\usepackage{multirow}
\usepackage[table]{xcolor}
\definecolor{mygray}{gray}{0.90}
\usepackage{subcaption}
\usepackage{caption}
\usepackage{booktabs}

\usepackage{stmaryrd}
\usepackage{marvosym}

\definecolor{mygray}{gray}{0.90}

\usepackage{hyperref} 
\hypersetup{
    pdftitle={Multi-task Neural Networks for Pain Intensity Estimation using Electrocardiogram and Demographic Factors},
    pdfauthor={Stefanos Gkikas},
    pdfsubject={Your subject here},
    pdfkeywords={Pain Recognition, ECG, Machine Learning, Age, Gender},
    bookmarksnumbered=true,     
    bookmarksopen=true,         
    bookmarksopenlevel=1,       
    colorlinks=true,
    citecolor=blue, 
    linkcolor=blue,           
    pdfstartview=Fit,           
    pdfpagemode=UseOutlines,    % this is the option you were lookin for
    pdfpagelayout=TwoPageRight
}

\usepackage{color}

\begin{document}
\title{Multi-task Neural Networks for Pain Intensity Estimation using Electrocardiogram and Demographic Factors} 

\titlerunning{Multi-task Neural Networks for Pain Intensity Estimation}

%\thanks{Supported by organization x.}}
\author{Stefanos Gkikas\inst{1}\orcidID{0000-0002-4123-1302} \and
Chariklia Chatzaki\inst{1}\orcidID{0000-0003-1312-8401} \and
Manolis Tsiknakis\inst{1,2}\orcidID{0000-0001-8454-1450}
}
\authorrunning{S. Gkikas et al.}
\institute{Department of Electrical and Computer Engineering, Hellenic Mediterranean University, Estavromenos, 71410, Heraklion,Greece \and 
Institute of Computer Science, Foundation for Research \& Technology-Hellas, Estavromenos, Heraklion, 70013, Greece}
\maketitle              % typeset the header of the contribution
\begin{abstract}
Pain is a complex phenomenon which is manifested and expressed by patients in various forms. The immediate and objective recognition of it is a great of importance in order to attain a reliable and unbiased healthcare system. In this work, we elaborate electrocardiography signals revealing     
the existence of variations in pain perception among different demographic groups. We exploit this insight by introducing a novel multi-task neural network for automatic pain estimation utilizing the age and the gender information of  each individual, and show its advantages compared to other approaches. 
\end{abstract}

\section{Introduction}
Pain according to  Williams and Craig \cite{williams_craig_2016} is \textit{"a distressing experience associated with actual or potential tissue damage with sensory, emotional, cognitive and social components"}.
As a biological mechanism, pain facilitates the identification of harmful situations by the activation of primary sensory neurons releasing prostanoids molecules and growth factors in the spinal cord \cite{khalid_tubbs_2017}.    
The two main types of pain are acute and chronic, where their main difference is related to the duration; the pain is considered as acute when is present less than three months and probably accompanied with clear physiological damage, while chronic when persist beyond
the normal healing time \cite{turk_2001}.
Pain affects people in a major degree, provoking a plethora of daily life challenges, especially in chronic pain condition, which often leads to mental health problems e.g. anxiety, depression and sleep related problems \cite{martinez_picard_2017}. 
In addition, there are various collateral negative effects, associated with opioid and drug overuse, addiction and poor social behavior relationships \cite{khalid_tubbs_2017}.
Pain is a serious issue concerns the whole society, since the consequences of it constitute clinical, economic and social constraints \cite{dinakar_stillman_2016}. Especially in health care systems, more than 50\% of the patients in hospitals are experience painful conditions, requiring large resources of medical and nursing stuff \cite{cordell_2002}.  
An important body of research indicates discrepancies regarding to pain manifestation and sensation in people of different gender and age, which increases the complication of pain management. In the meta-analysis of Riley et al. \cite{Riley_robinson_1998} refer that females demonstrate an elevated sensitivity for a variety of pain stimulus, while in the psychological review research \cite{bartley_fillingim_2013} the author deduced that females perceive increased irritation and pain in more areas of the body than males. 
Furthermore, alterations in pain sensation, presented in the study \cite{hadjistavropoulos_craig_2002} among elderly and youth population as well.      

In clinical settings, self-report is the gold standard for the evaluation of presence and intensity of pain, by rating scales and questionnaires. Nevertheless, this process is high labor-intensive, and the constant patient supervision is impracticable. In addition, the pain assessment is further intricate and demanding regarding to patients with communicational limitations, mental deficiency, severe illness, or infants \cite{werner_hamadi_2014}. Sufficient and impartial pain assessment is required, in order to provide the essential medical management to those in pain and prevent additional heath problems.    
The estimation of pain is derived from the interpretation of behavioural and physiological responses;   
the behavioural responses includes facial expressions, body-head movements and vocalizations, while the physiological are the electrical flow generated from neurochemical activities which provoke the sympathetic nervous system, something that can be manifested and perceived in the physiological signals \cite{stewart_2013} e.g. electrocardiography (ECG), 
electromyography (EMG) and skin conductance response (GSR).

This study is founded on our previous work \cite{gkikas_chatzaki_2022} investigating the variations of pain manifestation among different demographic groups, exploiting ECG signals.     
Furthermore, we extend our study by adopting neural networks as the main machine learning model, and propose a novel multi-task learning (MTL) neural network which exploits the demographic information by estimating the age and the gender beyond the pain level, in order to develop an improved automatic pain estimation system. 
The remaining of this paper is organised as follows: in Section \ref{related_work} we present the related work, in Section \ref{methods} we describe the process of the feature extraction and the development of the neural network, Section \ref{experiments_results} presents the conducted experiments and findings, and finally Section \ref{conclusion} concludes the paper.

\section{Related Work}
\label{related_work}
An important number of published research studies, founded on the utilization of biosignals in order to analyse the pain manifestation \cite{wang_wei_2020}\cite{martinez_picard_2017}.
A major reason of the preference of biosignals instead of vision modalities e.g. facial expressions, is related to circumstances where subdued lightning, facial occlusions or even face absence occur, especially in clinical 
settings where the conditions are far from being perfect.
Furthermore,  in several occasions people exhibit exaggerated symptoms through facial expressions and body posture in order to elicit self-interest \cite{Rohling_binder_1995}.

Utilizing ECG, EMG and electrodermal activity (EDA), researchers in \cite{kachele_thiam_2016}, extracted various handcrafted features including skewness, standard deviation and QRS complexes, performing multi-level classification through a Random Forest (RF) classifier. Similarly,  Amirial et al. \cite{amirian_kachele_2016} utilized  ECG, EMG and EDA, where several features extracted from the time and frequency domain.  By developing a Radial Basis Function (RBF) Neural Network,  they achieved accepted results, while in \cite{martinez_picard_2018_b} the authors designed a Recurrent Neural Network (RNN), feed it with  the extracted R peaks and interbeat Intervals (IBIs) from ECG signals.
Regarding to multi-task learning approaches, Lopez-Martinez and Picard \cite{martinez_picard_2017} proposed a multi-task neural network utilizing ECG and EDA signals, where beyond the pain estimation the model predicts the identity of the person. In a follow-up study of the same authors \cite{martinez_picard_2017_2} extended their work by adding visual modality, and furthermore, clustering subjects into different profiles according to their physiological and behavioural responses.

Recently, plethora of Deep Learning (DL) methods have been adopted in automatic pain estimation field, since in some cases their results are superior than traditional feature engineering and classical machine learning. 
The research of Wang et al. \cite{wang_wei_2020} utilized EEG potentials and an Autoencoder (AE) encoding the raw data to a compressed format, and classified them with a  Logistic Regressor (LR) classifier, while 
Yu et al. \cite{yu_sun_zhu_2020} developed a framework which was consisted of five convolutional modules in order to analyse three classes of pain, namely no pain, moderate and severe based on EEG signals. 
Thiam et al. \cite{thiam_bellmann_kestler_2019} designed deep 1D Convolutional Neural Networks employing ECG, EMG and GSR, experimenting with unimodal approaches, as well as multimodal fusion techniques.
Interestingly, the authors in \cite{huang_feng_2022} proposed a framework to compute pseudo heart rate gain from videos through a 3D convolutional neural network (CNN). Utilizing the pseudo physiological modality achieved high performance in binary and multi-class classification setting.

However, to the best of our knowledge, extremely limited work has been conducted on the automatic pain estimation research taking into consideration demographic factors.
The authors in \cite{hinduja_canavan_2020}, employing numerous biosignals such as EDA, respiration rate and diastolic blood pressure, as well as facial action units, revealed alterations in pain sensation among males and females. 
Similar are the findings in \cite{subramaniam_dass_2021}, where the conducted experiments which founded on a hybrid CNN-LSTM model and the utilization of  ECG and EDA,  indicating variations between the gender.
Finally, our previous work \cite{gkikas_chatzaki_2022}, beyond the gender differences reveled even higher variations including the factor of age.

\section{Methods}
\label{methods}
This section describes the employed pain database, the signal processing algorithm and feature extraction method,  as well as the design of the multi-task neural network.

\subsection{Dataset details}
In this study we employed the publicly available \textit{BioVid Heat Pain Database}  \cite{biovid_2013}, which combines (1) frontal facial videos,  (2) electrocardiogram, (3) electromyogram and (4) skin conductance level, recorded from 87 subjects (44 males, 43 females, age 20-65).
Currently, the BioVid Heat Pain Database, is the only publicly available database which in-corporates the age and gender of the subjects.
By subjecting heat stimulus on the right arm by a thermode, data were collected since the pain threshold (the temperature for which the participant’s sensing changes from heat to pain) and pain tolerance (the temperature at which the pain becomes intolerable) for each subject were determined. 
Based on the specific thresholds, 5 pain intensities defined: No pain (NP), mild pain (P1), moderate pain (P2), severe pain (P3), very severe pain (P4). 
The subjects were stimulated 20 times for each intensity, thus generating 100 samples for every modality. 
The sampling frequency of ECG recordings is equal to 512 Hz.
In this work we employed the pre-processed with a Butterworth band-pass filter ECG samples (87x100=8700) from Part A of the \textit{BioVid}.

\subsection{ECG processing and feature extraction}
\label{feature_extraction}
\begin{figure}[!ht]
\centering
\begin{floatrow}
  
\ffigbox[\FBwidth]{\caption{The PQRST complex}
\label{qrs}}
{\includegraphics[scale=0.75]{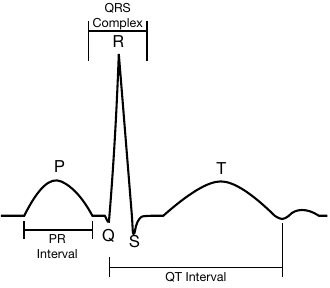} }\hspace{0.2cm}

\ffigbox[\FBwidth]{\caption{The flow diagram of the pre-processing procedure of the Pan-Tompkins algorithm \cite{gkikas_chatzaki_2022}}
\label{pan_tom}}
{\includegraphics[scale=0.86]{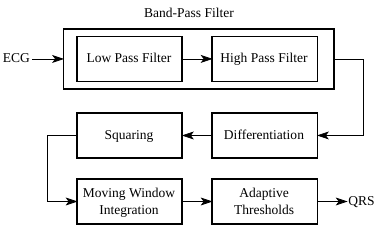} }

\end{floatrow}
\end{figure}

\begin{figure}
\centering
\includegraphics[scale=0.145]{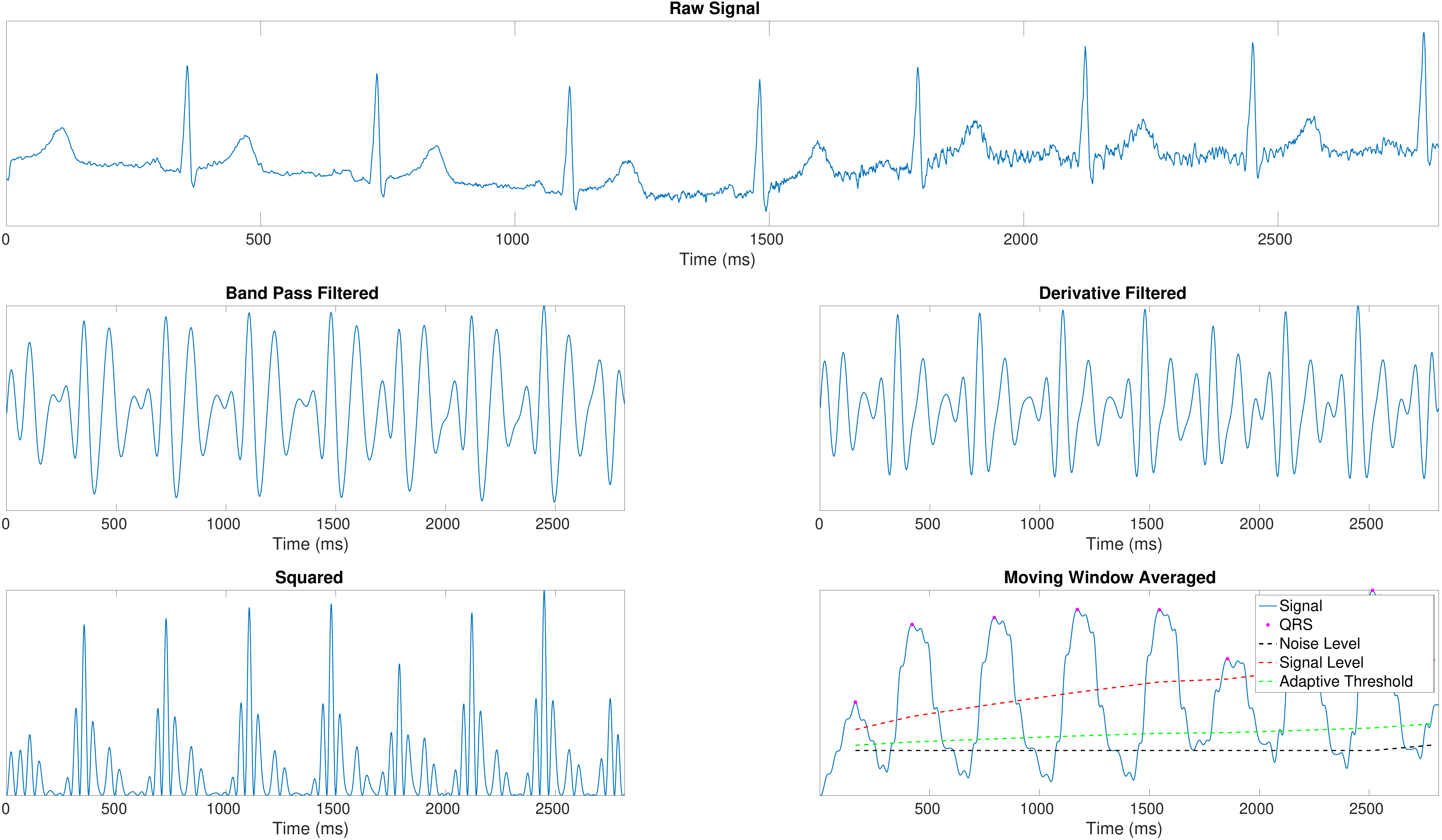}
\caption{ECG pre-processing with Pan-Tompkins algorithm \cite{gkikas_chatzaki_2022}}
\label{pan_tompkins_signal}
\end{figure}

An ECG signal reflects the electrical activity of the heart during time, where cardiac muscles depolarize and repolarize during a cardiac cycle. The cardiac cycle describes the undergoing activity from the beginning of one heartbeat to the beginning of the next, which in an ECG complex consists of a PQRST complex. The P wave indicates atrial depolarization, while the QRS complex represents ventricular depolarization and contraction. The T wave describes repolarization of ventricles. Consequently, the ECG analysis prerequisites the decomposition of the PQRST complex (see Fig. \ref{qrs}). By the accurate detection of R wave, which is the most prevalent peak in the complex, we are capable to calculate the heart rate (HR) and the heart rate variability (HRV), which is related with the time interval between consecutive R waves, called as RR interval or Interbeat interval. In this study, we adopt the Pan-Tompkins Algorithm \cite{pan_tompkins_1985} for the QRS detection. The specific algorithm is widely used and evaluated over the years, with the results supporting its efficiency even in noisy and low-quality data \cite{fariha_ikeura_2020}. The synthesis of the algorithm emerges in two stages: the pre-processing and the decision; the pre-processing deals with removing noise and artifacts, as well as smoothing the signal and increasing the QRS slope, while the decision, encompass the initial QRS detection based on adaptive thresholds, a search back for missed QRS complexes, and a process for T wave discrimination. 
The basic flow diagram of Pan-Tompkins algorithm as well as the pre-process procedure applied in raw ECG presented in Fig. \ref{pan_tom} and Fig. \ref{pan_tompkins_signal} respectively.

%%%%%%%%%%%%%%%%%%%%%%%%%%%%%%%%%%%%%%%%%%%%%%%%%%%%%%%%%%%%%%%%%%%%%%%%%%%%%%%%%%%%%%%%%%%%%
%%%%%%%%%%%%%%%%%%%%%%%%%%%%%%%%%%%%%%%%%%%%%%%%%%%%%%%%%%%%%%%%%%%%%%%%%%%%%%%%%%%%%%%%%%
Following the accurate detection of R waves, the inter-beat intervals (IBIs) were estimated, and the most important relevant features were extracted. Particularly, the mean of IBIs, the root mean square of successive differences (RMSSD), the standard deviation of IBIs (SDNN), the slope of the linear regression of IBIs, the ratio of SDNN to RMSSD, and the heartbeat rate, were calculated as in detail described in our previous work \cite{gkikas_chatzaki_2022}.

%%%%%%%%%%%%%%%%%%%%%%%%%%%%%%%%%%%%%%%%%%%%%%%%%%%%%%%%%%%%%%%%%%%%%%%%%%%%%%%%%%%%%%%
%\begin{enumerate}
%\item Mean of IBIs \cite{gkikas_chatzaki_2022}
%\begin{equation} 
%\mu=\dfrac{1}{n}\sum_{i=1}^{n}(RR_{i+1}-RR_{i})
%\end{equation}
%where $RR$ are consecutive $R$ peaks. 
%
%\item Root mean square of successive differences \cite{gkikas_chatzaki_2022}
%\begin{equation} 
%RMSSD=\sqrt{  \dfrac{1}{n-1}   \sum_{i=1}^{n-1}(RR_{i+1}-RR_{i})^{2}  }
%\end{equation}
%\item Standard deviation of IBIs \cite{gkikas_chatzaki_2022}
%\begin{equation} 
%SDNN=\sqrt{  \dfrac{1}{n-1}   \sum_{i=1}^{n}(RR_{i}-\mu)^{2} }
%\end{equation}
%\item Slope of the linear regression of IBIs \cite{gkikas_chatzaki_2022}
%\begin{equation}
%{A}^{T}{A}x=A^{T}{b}
%\end{equation}
%based on the least-square approximation, where $b$ is the vector of $RR$ peak intervals and $A$ is the corresponding time series.
%\item Ratio of SDNN to RMSSD \cite{gkikas_chatzaki_2022}
%\begin{equation}
%SR=\dfrac{SDNN}{RMSSD}
%\end{equation}
%\item Heart beat rate \cite{gkikas_chatzaki_2022}
%\begin{equation}
%HR=\dfrac{60\cdot FS}{\mu}
%\end{equation}
%where $FS$ is the frequency of ECG recording and is equal to 512 Hz. Figure \ref{pan_tompkins_signal} shows the raw ECG signal and the applied algorithm's steps as well.  
%\end{enumerate}
%%%%%%%%%%%%%%%%%%%%%%%%%%%%%%%%%%%%%%%%%%%%%%%%%%%%%%%%%%%%%%%%%%%%%%%%%%%%%%%%%%

%%%%%%%%%%%%%%%%%%%%%%%%%%%%%%%%%%%%%%%%%%%%%%%%%%%%%%%%%%%%%%%%%%%%%%%%%%%%%%%%

\subsection{Neural network}
Our proposed neural network trained in two different settings; with single-task learning (STL) and multi-task learning (MTL), where the latter beyond the pain estimation it involves the simultaneous training of age and/or gender estimation.

\subsubsection{Single-task neural network}
\label{st-nn}
The proposed neural network consists of two sub-networks; the encoder which is mapping the original feature vectors into a higher dimensional space, and the task-specific classifier.
Our method employs fully-connected (FC) layers for both the encoder and the classifier, each one defined as follows:

\begin{equation}
z_{i}(s)=b_{i}+\sum_{j=1}^{n_{in}}W_{ij}s_{j} \qquad \textrm{for} \quad i=1,..,n_{out}
\end{equation}
where $z_{i}$ is the outcome of the linear aggregation of incoming inputs $s_{j}$, and each input is weighted by $W_{ij}$ and biased by $b_{i}$. Every layer of the encoder is followed by a nonlinear activation function, namely rectified linear unit (ReLU) defined as:
\begin{equation}
\sigma(z)=
	\begin{cases}
	1, \quad z\geq0 \\
	0, \quad z<0
	\end{cases}
\end{equation}
while the classifier' layers are connected without nonlinearity. The encoder consists of 4 FC layers with 256, 512, 1024 and 1024 neurons respectively, while the classifier consists of 2 layers with 1024 and $n$ neurons, where $n$ is the number of the corresponding pain classes. In table \ref{table:parameters}, we list the hyper-parameters of our network.

%%%%%%%%%%%%%%%%%%%%%%%%%%%%%%%%%%%%%%%%%%%%%%%%%%%%%%%%%%%%%%%%%%%%%%%%%%%%%%%%%%%%%%%%%%%%%%%%
\begin{table}
\renewcommand{\arraystretch}{1.5}
\begin{threeparttable}
    \caption{Training hyper-parameters used in our method.} 
     \label{table:parameters}

\begin{tabular}{P{1.2cm} |P{1.6cm}| P{1.6cm} |P{1.1cm} |P{1.4cm} |P{1.5cm} |P{1.4cm} |P{1.1cm}}

\toprule
Epochs  &Optimizer  &Learning rate &LR decay &Weight decay &Warmup epochs &Label smooth &EMA\\
\hline
\hline
300 &AdamW &1e-3 &cosine &0.1 &50 &0.1 &$\checkmark$\\
\bottomrule

\end{tabular}
\end{threeparttable}
\end{table}
%%%%%%%%%%%%%%%%%%%%%%%%%%%%%%%%%%%%%%%%%%%%%%%%%%%%%%%%%%%%%%%%%%%%%%%%%%%%%%%%%%%%%%%%%%%%%%%%%

\subsubsection{Multi-task neural network}
\label{mt-nn}
\begin{figure}
\centering
\includegraphics[scale=0.76]{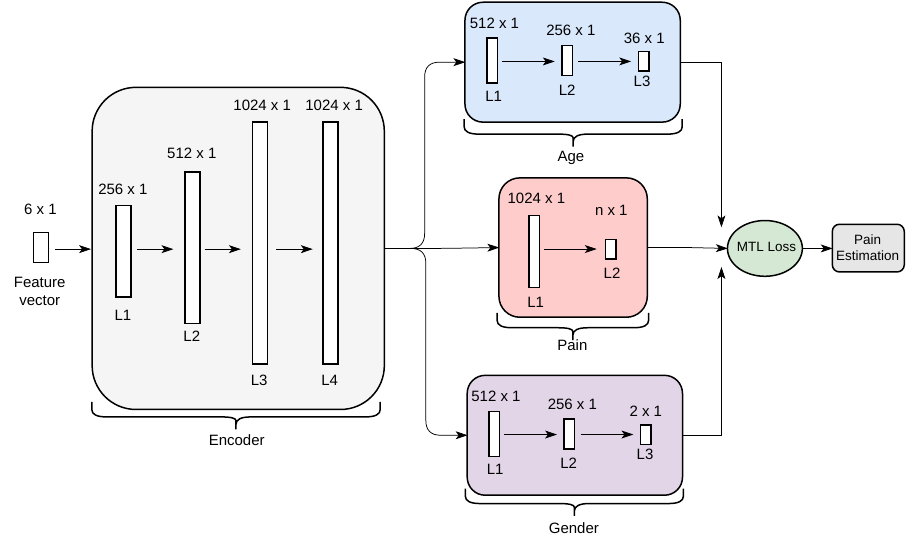}
\caption{Illustration of the MTL network. The output vectors' size of the network is: for Pain classifier $nx1$ where n the number of pain estimation tasks (i.e. 2 for binary classification, 5 for multi-class classification), for Age classifier $36x1$ where 36 is equal to the number of possible values of subjects' age, for Gender classifier $2x1$ where 2 the number of possible values (i.e males, females)}
\label{mtl}
\end{figure}

The specific machine learning method is founded on the principle of sharing representations between related tasks, 
enabling the model to generalize better on the original task, i.e. pain estimation. In this settings, we retained the identical encoder and pain classifier, while we added two auxiliary networks, for age and gender estimation respectively. 
Fig. \ref{mtl} presents the architecture of proposed MTL neural network. 
The objective of the network is the minimization of the three losses, simultaneous. We adopt and extend the proposal of \cite{kendall_2018} regarding the multi-task learning loss, in which learned weights multiple the loss functions by considering the homoscedastic uncertainty of each task:

\begin{equation}
L_{total}= [e^{w1}L_{Pain}+w_{1}]c_{1} + [e^{w2}L_{Age}+w_{2}]c_{2} + [e^{w3}L_{Gender}+w_{3}]c_{3}
\end{equation}
where $L$ is the corresponding loss, $w$ the weights and $c$ the coefficients which restrain the $L_{Age}$ and $L_{Gender}$
in order to influence the learning procedure into the pain estimation task. We mention that all the tasks addressed as classification problems, adopting the \textit{cross-entropy} loss with label smoothing:
\begin{equation}
L_{D} = -\sum_{\delta \in D}\sum^{n_{out}}_{i=1}p(i|x_\delta)\log[q(i|x_\delta)]
\end{equation}
where $D$ is the pain database, $p(i|x_\delta)=1-\epsilon$ and $p(i\neq i_\delta|x_\delta)=\epsilon/(n_{out}-1)$ is the distribution over the components $i$ of the ground truth,
and $q(i|x_\delta)$ is the distribution over the components $i$ of the networks' output.

\section{Experiments and results}
\label{experiments_results}
\subsection{Demographic groups}
Utilizing the single-task neural network (ST-NN) we conducted the first body of experiments related to the influence of demographic factors. Specifically, adopting the idea of \cite{gkikas_chatzaki_2022} we developed five schemes; (1) basic scheme, utilizing all the subjects of the database, (2) gender scheme, dividing the subjects into males and females, (3) age scheme, based on the subjects' age forming three groups i.e. \textit{20-35}, \textit{36-50}, \textit{51-65}, and (4) gender-age scheme combining the gender and age of the subjects, creating six groups i.e. \textit{Males 20-35}, \textit{Females 20-35}, \textit{Males 36-50}, \textit{Females 36-50}, \textit{Males 51-65}, \textit{Females 51-65}. We note that all the experiments of this study conducted in binary and multi-class classification settings. In particular: (1) NP vs P1, (2) NP vs P2, (3) NP vs P3, (4) NP vs P4 regarding the binary classification and finally (5) multi-class pain classification, utilizing all the available pain classes of the database. 

\begin{figure}
     \centering
     \begin{subfigure}{\textwidth}
         \centering
         \includegraphics[scale=0.40]{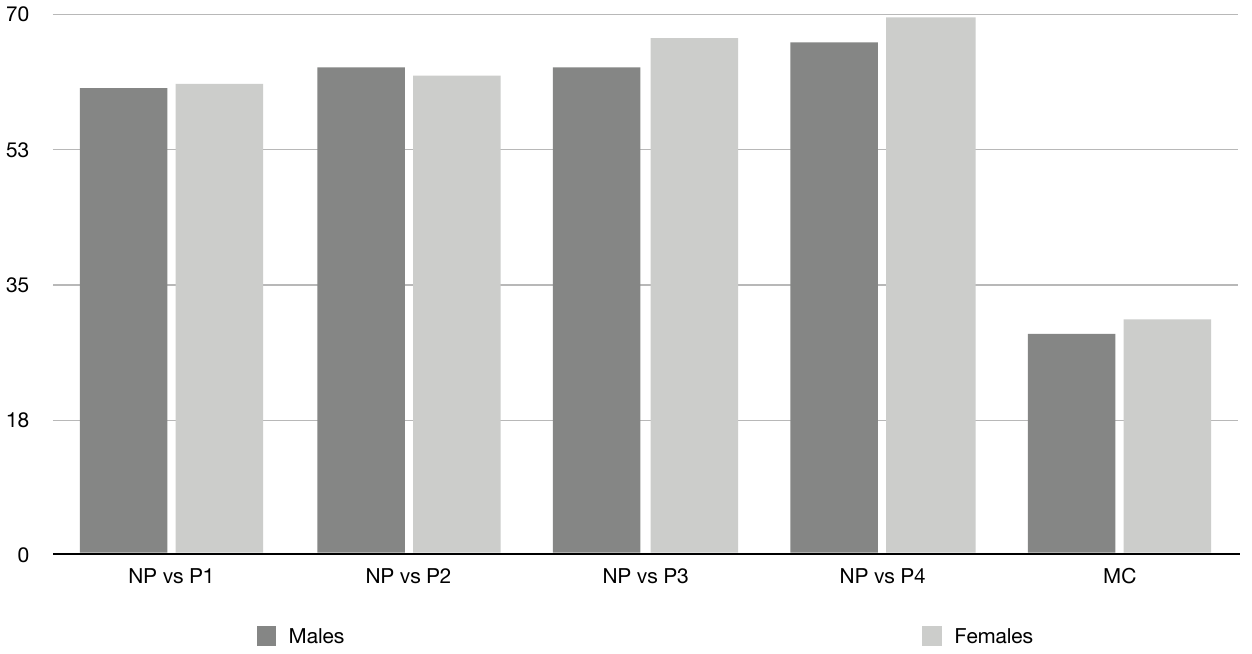}
         \caption{Gender}
         \label{gender}
     \end{subfigure}
     \par\bigskip

     \begin{subfigure}{\textwidth}
         \centering
         \includegraphics[scale=0.40]{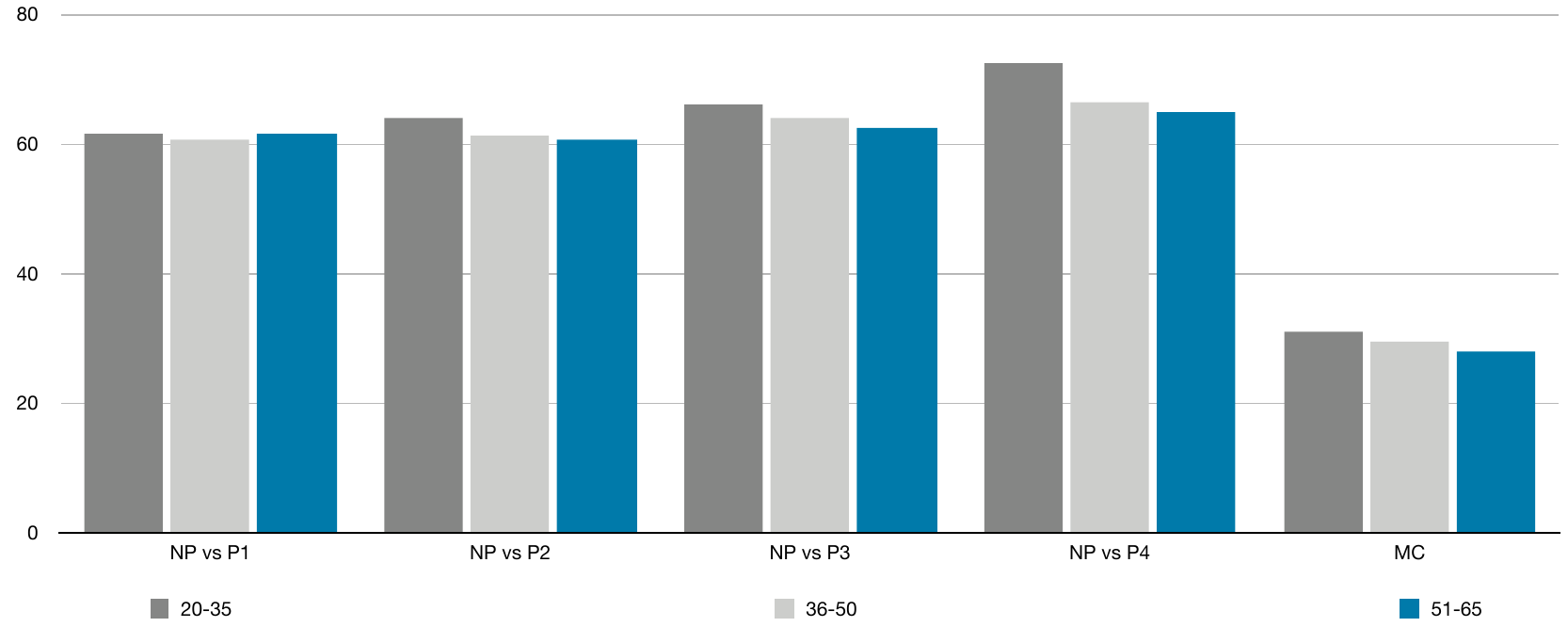}
         \caption{Age}
         \label{age}
     \end{subfigure}
     \par\bigskip

     \begin{subfigure}{\textwidth}
         \centering
         \includegraphics[scale=0.40]{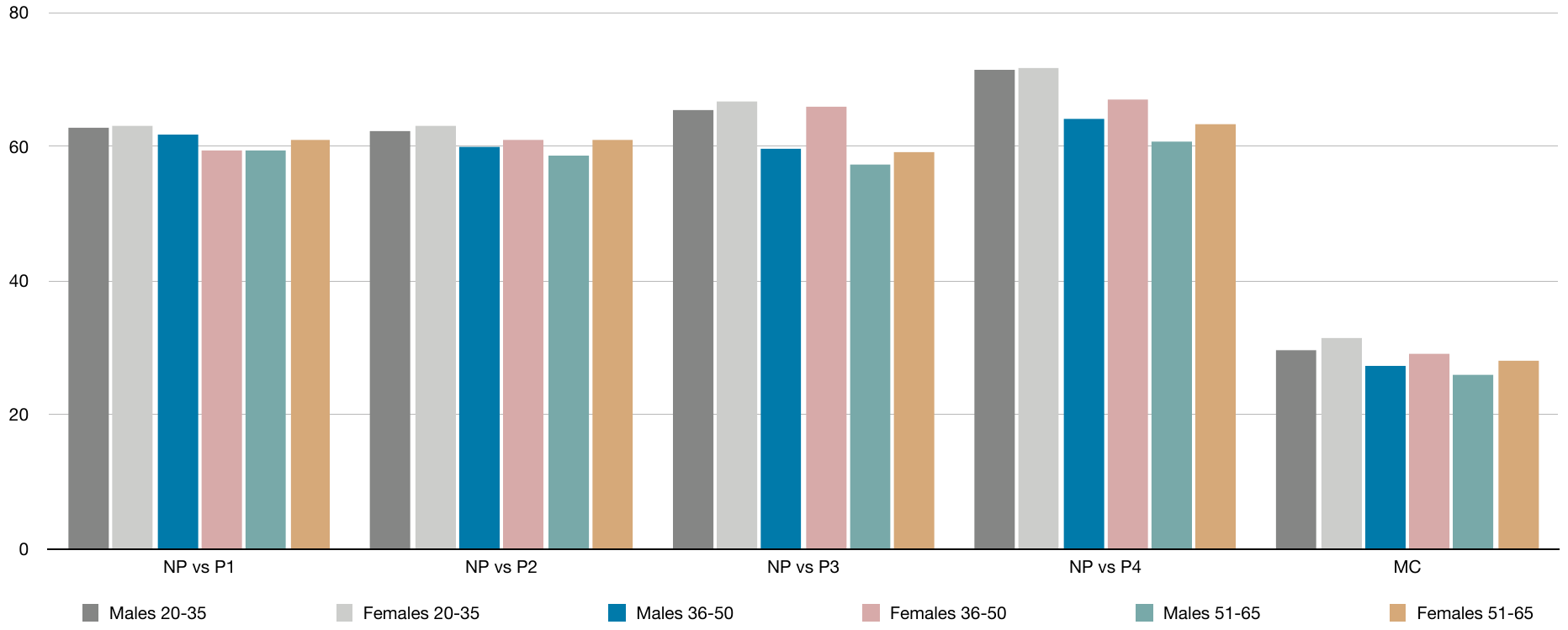}
         \caption{Gender-Age}
         \label{gender_age}
     \end{subfigure}

\caption{Classification results on different Schemes}
\label{schemes}
\end{figure}

Table \ref{table:basic} presents the classification results of Basic scheme, utilizing all the subjects of the database. 
For the multi-class pain classification we achieved 29.43\%, while the accuracy on NP vs P1 attained 61.15\% and reaching up to 68.82\% on NP vs P4. We observe that as the pain intensity raise, the performance scores increased as well, revealing the challenges to recognise the low magnitude of pain severity.       
According to the gender scheme (see Table \ref{table:gender}) there are clear differences among males and females, especially in higher pain intensities.
Specifically, in NP vs P4 females attained 69.48\% over 66.48\%, while the two genders present 1.63\% variance in total, exhibiting that females characterized by a higher pain sensitivity.    
Figure \ref{gender} outlines the classification differences among the gender.  
On the age scheme (see Table \ref{table:age}), and particularly on NP vs P4 the group \textit{20-35} presents 72.58\% over 66.29\% and 64.91\% from the groups \textit{36-50} and \textit{51-65} respectively, while in lower pain intensities the differences are less noticeable. Nevertheless, the specific scheme reveals that the factor of age influences the pain manifestation, especially for older population. Figure \ref{age} depicts the results of age scheme. 

In the last scheme, dividing the subjects into additional number of groups, we are able to study them in a more precisely manner, and obtain and better understanding about the correlation of gender and age with the pain sensation. In Table \ref{table:gender_age} we observe that in task NP vs P4 the group \textit{Females 20-35} achieved the highest accuracy with 71.67\%, exceeding the group \textit{Males 51-65} by 11\% which obtained the minimal performance, and described as the least sensitive group. 
Similar are the observations on the multi-class classification, as well as in the remaining pain tasks, where
\textit{Females 20-35} and \textit{Males 51-65} presented the uppermost and the minor pain estimation accuracies respectively. 
This corroborates that females defined with elevated pain manifestation, while males, especially seniors possess diminished sensation. We mention that in some cases e.g. \textit{Males 20-35}, \textit{Males 36-50} despite the fact the pain intention in increased, the classification accuracy is diminished, something also observed in \cite{gkikas_chatzaki_2022}. 
An possible interpretation would be the accustomation of the subjects in pain situations, especially in low intensities. Figure \ref{gender_age} visualize the performances of the gender-age scheme.

%%%%%%%%%%%%%%%%%%%%%%%%%%%%%%%%%%%%%%%%%%%%%%%%%%%%%%%%%%%%%%%%%%%%%%%%%%%%%%%%%%%%%%%%%%%%%%%%%%%%%
\renewcommand{\arraystretch}{1.1}
\begin{table}
\begin{threeparttable}
    \caption{Classification results of the Basic Scheme, reported on \% accuracy.}
     \label{table:basic}
     
\begin{tabular}{ P{1.0cm} |P{1.7cm}| P{1.5cm}| P{1.5cm} |P{1.5cm} |P{1.5cm} |P{1.5cm}}
\hline

\multirow{2}[3]{*}{Group}  &\multirow{2}[3]{*}{Algorithm} &\multicolumn{5}{c}{Task}\\
\cmidrule(lr){3-7}
\ & &NP vs P1  &NP vs P2 &NP vs P3 &NP vs P4 &MC\\

\hline\hline
All &ST-NN &61.15 &62.87 &65.14 &68.82 &29.43 \\
\hline
\end{tabular}

         \begin{tablenotes}
          \scriptsize
           \item ST-NN: single-task neural network \space NP: no pain \space P1: mild pain \space P2: moderate pain \space P3: severe pain \space P4: very severe pain \space MC: multi-classification 
          \end{tablenotes}
\end{threeparttable}
\end{table}
%%%%%%%%%%%%%%%%%%%%%%%%%%%%%%%%%%%%%%%%%%%%%%%%%%%%%%%%%%%%%%%%%%%%%%%%%%%%%%%%%%%%%%%%%%%%%%%%%%%

\renewcommand{\arraystretch}{1.1}
\begin{table}
\begin{threeparttable}
    \caption{Classification results of the Gender Scheme, reported on \% accuracy.}
     \label{table:gender}
     
\begin{tabular}{ P{1.3cm} |P{1.7cm}| P{1.5cm}| P{1.5cm} |P{1.5cm} |P{1.5cm} |P{1.5cm}}
\hline

\multirow{2}[3]{*}{Group}  &\multirow{2}[3]{*}{Algorithm} &\multicolumn{5}{c}{Task}\\
\cmidrule(lr){3-7}
\ & &NP vs P1  &NP vs P2 &NP vs P3 &NP vs P4 &MC\\

\hline\hline
Males &ST-NN &60.40 &63.24 &63.18 &66.48 &28.61\\
Females &ST-NN &60.87 &62.15 &66.98 &69.48 &30.59\\
\hline
\end{tabular}
\end{threeparttable}
\end{table}

%%%%%%%%%%%%%%%%%%%%%%%%%%%%%%%%%%%%%%%%%%%%%%%%%%%%%%%%%%%%%%%%%%%%%%%%%%%%%%%%%%%%%%%%%%%%%%%%%%%

\renewcommand{\arraystretch}{1.1}
\begin{table}
\begin{threeparttable}
    \caption{Classification results of the Age Scheme, reported on \% accuracy.}
     \label{table:age}
     
\begin{tabular}{ P{1.3cm} |P{1.7cm}| P{1.5cm}| P{1.5cm} |P{1.5cm} |P{1.5cm} |P{1.5cm}}
\hline

\multirow{2}[3]{*}{Group}  &\multirow{2}[3]{*}{Algorithm} &\multicolumn{5}{c}{Task}\\
\cmidrule(lr){3-7}
\ & &NP vs P1  &NP vs P2 &NP vs P3 &NP vs P4 &MC\\

\hline\hline
20-35 &ST-NN &61.58 &64.08 &66.08 &72.58 &31.07\\
36-50 &ST-NN &60.52 &61.38 &64.05 &66.29 &29.59\\
51-65 &ST-NN &61.70 &60.80 &62.50 &64.91 &27.82\\
\hline
\end{tabular}
\end{threeparttable}
\end{table}
%%%%%%%%%%%%%%%%%%%%%%%%%%%%%%%%%%%%%%%%%%%%%%%%%%%%%%%%%%%%%%%%%%%%%%%%%%%%%%%%%%%%%%%%%%%%%%%%%%%
\renewcommand{\arraystretch}{1.1}
\begin{table}
\begin{threeparttable}
    \caption{Classification results of the Gender-Age Scheme, reported on \% accuracy.}
     \label{table:gender_age}
     
\begin{tabular}{ P{2.0cm} |P{1.7cm}| P{1.5cm}| P{1.5cm} |P{1.5cm} |P{1.5cm} |P{1.5cm}}
\hline

\multirow{2}[3]{*}{Group}  &\multirow{2}[3]{*}{Algorithm} &\multicolumn{5}{c}{Task}\\
\cmidrule(lr){3-7}
\ & &NP vs P1  &NP vs P2 &NP vs P3 &NP vs P4 &MC\\

\hline\hline
Males   20-35 &ST-NN &62.83 &62.33 &65.5 &71.33 &29.73\\
Males   36-50 &ST-NN &61.79 &60.00 &59.64 &64.11 &27.14\\
Males   51-65 &ST-NN &59.50 &58.67 &57.33 &60.67 &26.07\\
Females 20-35 &ST-NN &63.17 &63.17 &66.83 &71.67 &31.53\\
Females 36-50 &ST-NN &59.50 &61.00 &65.83 &67.00 &29.13\\
Females 51-65 &ST-NN &60.96 &60.96 &59.23 &63.27 &27.69\\
\hline
\end{tabular}
\end{threeparttable}
\end{table}
%%%%%%%%%%%%%%%%%%%%%%%%%%%%%%%%%%%%%%%%%%%%%%%%%%%%%%%%%%%%%%%%%%%%%%%%%%%%%%%%%%%%%%%%%%%%%%%%%%%

%%%%%%%%%%%%%%%%%%%%%%%%%%%%%%%%%%%%%%%%%%%%%%%%%%%%%%%%%%%%%%%%%%%%%%%%%%%%%%%%%%%%%%%%%%%%%%%%%%%
\pagebreak
\subsection{Augmentation of feature vectors}
Based on the observations from the previous section regarding the influence of demographic factors in pain manifestation, we 
investigate the functional exploitation of subjects' demographic information.
We conducted a set of experiments utilizing the ST-NN and the feature vectors which augmented by the expansion of them with demographic elements.
Specifically, the original feature vectors, consisting of six features (see \ref{feature_extraction}), expanding by an additional feature (i.e. the subjects' gender or age), or two additional features (the subjects' gender and age). 
In this manner, utilizing the new set of features we carried out the identical experiments related to pain estimation tasks.
Table \ref{table:augm} presents the classification results, where we observe increased performances adopting the approach of augmented feature vectors.
In particular, the most effected type of augmentation is the combination of Gender and Age feature, which improved the mean pain estimation performance by 0.55\%, while the utilization of them individually, enhanced the classification accuracy, but in a lower degree.      

%%%%%%%%%%%%%%%%%%%%%%%%%%%%%%%%%%%%%%%%%%%%%%%%%%%%%%%%%%%%%%%%%%%%%%%%%%%%%%%%%%%%%%%%%%%%%%%%%%%%%
\renewcommand{\arraystretch}{1.1}
\begin{table}
\begin{threeparttable}
    \caption{Comparison of classification results adopting the feature augmentation approach, reported on \% accuracy.}
     \label{table:augm}
     
\begin{tabular}{ P{1.0cm} |P{1.7cm}| P{1.1cm} |P{1.5cm}| P{1.5cm} |P{1.5cm} |P{1.5cm} |P{1.4cm}}
\hline

\multirow{2}[3]{*}{Group}  &\multirow{2}[3]{*}{Algorithm} &\multirow{2}[3]{*}{Aux.} &\multicolumn{5}{c}{Task}\\
\cmidrule(lr){4-8}
\ & & &NP vs P1  &NP vs P2 &NP vs P3 &NP vs P4 &MC\\

\hline\hline
All &ST-NN &- &61.15 &62.87 &65.14 &68.82 &29.43 \\
\hline
All &ST-NN &F(G)  &61.44 &63.19 &65.00 &68.79 &29.68\\
All &ST-NN &F(A)  &61.21 &62.67 &65.66 &69.57 &29.71\\
All &ST-NN &F(GA) &61.09 &63.48 &66.21 &69.54 &29.86\\
\hline
\end{tabular}

         \begin{tablenotes}
          \scriptsize
           \item Aux: Auxiliary information \space -: original feature vectors \space F(G): feature vectors with the additional feature of gender \space F(A): feature vectors with the additional feature of age \space F(GA): feature vectors with the additional features of gender and age
          \end{tablenotes}
\end{threeparttable}
\end{table}
%%%%%%%%%%%%%%%%%%%%%%%%%%%%%%%%%%%%%%%%%%%%%%%%%%%%%%%%%%%%%%%%%%%%%%%%%%%%%%%%%%%%%%%%%%%%%%%%%%% 

\subsection{Multi-Task Neural Network} The final set of experiments conducted in a multi-task learning manner, utilizing the proposed MT-NN described in \ref{mt-nn}.
The classification performances of MT-NN with the additional tasks of (1) gender estimation, (2) age
estimation and (3) gender \& age estimation simultaneously, are presented in Table \ref{table:mtl}.
The results of the previous approaches based on the ST-NN method, are presented as well in Table \ref{table:mtl}.
We observe that the additional task of gender estimation performed inferior compared to others tasks, while the combination of gender \& age achieved the highest performance in four of the five tasks. Specifically, in the multi-class classification attained 30.24\%, while in NP vs P1 62.82\% which are the greatest results compared to every presented method in this study.
Similarly, in NP vs P3 and NP vs P4 outperformed the gender and age additional tasks, however underperformed to ST-NN approaches with the augmented features.    
Finally, in NP vs P2 the additional task of age estimation performed superior achieving 63.97\%, which is also the highest attained result in this study.

\begin{figure}
     \centering
     \begin{subfigure}{\textwidth}
         \centering
         \includegraphics[scale=0.45]{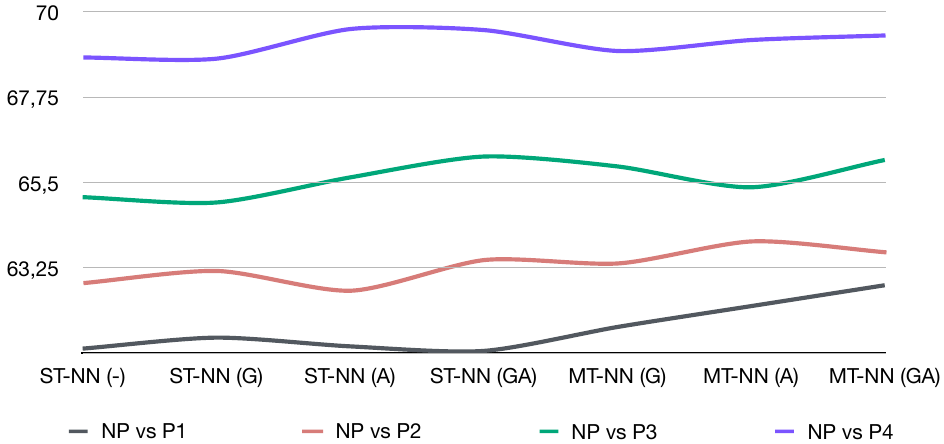}
         \caption{Binary classification}
         \label{np}
     \end{subfigure}
     \par\bigskip
     
     \begin{subfigure}{\textwidth}
         \centering
         \includegraphics[scale=0.45]{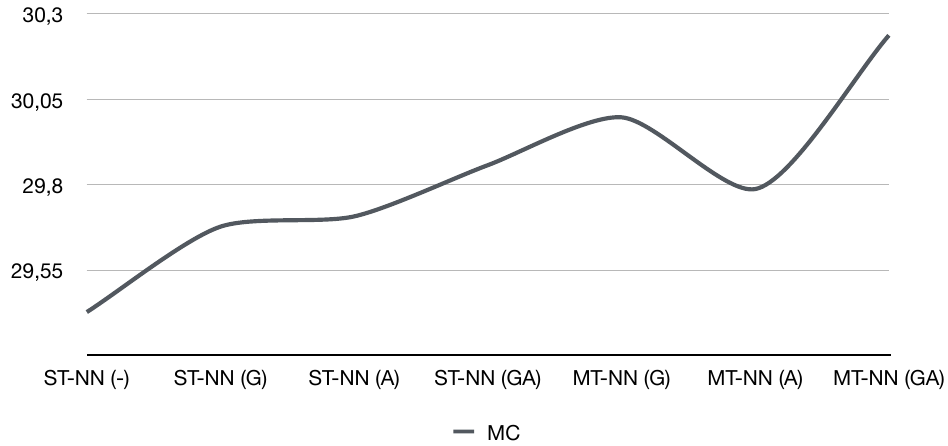}
         \caption{Multi-class classification}
         \label{mc}
     \end{subfigure}

\caption{Comparison of classification performances adopting different neural networks approaches}
\label{all_nns}
\end{figure}

Regarding the overall achieved performances of MT-NN compared to the ST-NN approaches (i.e. original features vectors and augmented feature vectors), we observe an increase of 0.71\% and 0.39\% respectively, in relation to     
the mean pain estimation accuracy of all tasks. Figure \ref{all_nns} illustrate the comparison of every neural network approach  
presented in this study, according to multi-class and binary classification tasks as well.

%%%%%%%%%%%%%%%%%%%%%%%%%%%%%%%%%%%%%%%%%%%%%%%%%%%%%%%%%%%%%%%%%%%%%%%%%%%%%%%%%%%%%%%%%%%%%%%%%%%%%
\renewcommand{\arraystretch}{1.1}
\begin{table}
\begin{threeparttable}
    \caption{Comparison of classification results adopting the MT-NN approach, reported on \% accuracy.}
     \label{table:mtl}
     
\begin{tabular}{ P{1.0cm} |P{1.7cm}| P{1.1cm} |P{1.5cm}| P{1.5cm} |P{1.5cm} |P{1.5cm} |P{1.4cm}}
\hline

\multirow{2}[3]{*}{Group}  &\multirow{2}[3]{*}{Algorithm} &\multirow{2}[3]{*}{Aux.} &\multicolumn{5}{c}{Task}\\
\cmidrule(lr){4-8}
\ & & &NP vs P1  &NP vs P2 &NP vs P3 &NP vs P4 &MC\\

\hline\hline
All &ST-NN &- &61.15 &62.87 &65.14 &68.82 &29.43 \\
\hline
All &ST-NN &F(G)  &61.44 &63.19 &65.00 &68.79 &29.68\\
All &ST-NN &F(A)  &61.21 &62.67 &65.66 &69.57 &29.71\\
All &ST-NN &F(GA) &61.09 &63.48 &66.21 &69.54 &29.86\\
\hline
All &MT-NN &T(G)  &61.72 &63.39 &65.95 &68.99 &30.00\\
All &MT-NN &T(A)  &62.72 &63.97 &65.40 &69.28 &29.79\\
All &MT-NN &T(GA) &62.82 &63.68 &66.12 &69.40 &30.24\\
\end{tabular}

         \begin{tablenotes}
          \scriptsize
           \item T(G): MT-NN with the additional task of gender estimation \space T(A): MT-NN with the additional task of age estimation \space T(GA): MT-NN with the additional task of gender and age estimation  
          \end{tablenotes}
\end{threeparttable}
\end{table}

%%%%%%%%%%%%%%%%%%%%%%%%%%%%%%%%%%%%%%%%%%%%%%%%%%%%%%%%%%%%%%%%%%%%%%%%%%%%
%%%%%%%%%%%%%%%%%%%%%%%%%%%%%%%%%%%%%%%%%%%%%%%%%%%%%%%%%%%%%%%%%%%%%%%%%%%%

\subsection{Comparison with existing methods}
Finally, in this section we compare our accomplished results utilizing the MT-NN employed the additional tasks of gender and age estimation, with corresponding studies which utilized the electrocardiography signals of the Part A of \textit{BioVid} database with all 87 participants, and followed the same evaluation protocol i.e. leave-one-subject-out (LOSO) cross validation for the purpose of an equitable comparison. The results depicted in Table \ref{table:comparison}, including studies utilizing hand-crafted features and classic machine learning algorithms \cite{gkikas_chatzaki_2022}\cite{werner_hamadi_2014}, 
\textit{end-to-end} deep learning approaches \cite{huang_feng_2022}\cite{thiam_bellmann_kestler_2019}, and methods combining hand-crafted features with deep learning classification algorithms \cite{martinez_picard_2018_b}. 
Our method exploiting the carefully designed ECG features, followed by the high-dimensional mapping from the encoder with the combination of the multi-task learning neural networks, was able to outperform other approaches in every pain estimation task, either in binary or multi-class classification setting.

%%%%%%%%%%%%%%%%%%%%%%%%%%%%%%%%%%%%%%%%%%%%%%%%%%%%%%%%%%%%%%%%%%%%%%%%%%%%
\renewcommand{\arraystretch}{1.1}
\begin{table}
\begin{threeparttable}
    \caption{Comparison of studies which utilized \textit{BioVid}, ECG signals and LOSO cross validation}
     \label{table:comparison}
     
\begin{tabular}{ P{3.8cm} |P{1.5cm}| P{1.5cm} |P{1.5cm} |P{1.5cm} |P{1.4cm}}
\hline

\multirow{2}[3]{*}{Study}  &\multicolumn{5}{c}{Task}\\
\cmidrule(lr){2-6}
\ &NP vs P1  &NP vs P2 &NP vs P3 &NP vs P4 &MC\\

\hline\hline
Gkikas et al. \cite{gkikas_chatzaki_2022}$^\dagger$ &52.38 &52.78 &55.37 &58.62 &23.79\\
Huang et al. \cite{huang_feng_2022}$^{\star\odot}$ &- &- &- &65.00 &28.50\\
Martinez and Picard \cite{martinez_picard_2018_b}$^\divideontimes$ &- &- &- &57.69 &-\\
Thiam et al. \cite{thiam_bellmann_kestler_2019}$^\star$ &49.71 &50.72 &52.87 &57.04 &23.23\\
Wernel et al. \cite{werner_hamadi_2014}$^\dagger$ &48.70 &51.60 &56.50 &62.00 &-\\
This study$^\divideontimes$ &\cellcolor{mygray}62.82 &\cellcolor{mygray}63.68 &\cellcolor{mygray}66.12 &\cellcolor{mygray}69.40 &\cellcolor{mygray}30.24\\

\end{tabular}

         \begin{tablenotes}
          \scriptsize
           \item $^\dagger$:hand-crafted features and classic machine learning \space $^\star$: end-to-end deep learning \space 					 $^\divideontimes$: hand-crafted features with deep learning classification algorithms \space $^\odot$: pseudo heart rate gain extracted from visual modality
          \end{tablenotes}
\end{threeparttable}
\end{table}
%%%%%%%%%%%%%%%%%%%%%%%%%%%%%%%%%%%%%%%%%%%%%%%%%%%%%%%%%%%%%%%%%%%%%%%%%%%%%%%
\section{Conclusion}
\label{conclusion}
This work explored the application of multi-task learning neural networks for automatic pain estimation based on  electrocardiography signals. Utilizing the Pan-Tompkins algorithm for the detection of QRS complexes we extracted essential features related to IBIs. Several experiments conducted in order to investigate the relation of gender and age with the pain manifestation, revealing the great influence of them to pain perception. Furthermore, we proposed two methods in order to exploit the demographic information enhancing the pain estimation results. First, the original feature vectors augmented with the subjects' demographic elements which indeed improved the classification accuracy. Second, the multi-task learning neural network integrating the pain estimation task with the gender and age estimation as well. The latter demonstrated superior results compared to previous presented methods in this paper, as well as to other related studies. 
This suggests that extracted domain specific features combined with properly designed deep learning architectures and demographic factors are able to accomplish great results.

In summary, we suggest that clinical pain management tools need to be designed in such a way so that patients' demographic information, which clearly influence pain manifestation, are  included. However, from the machine learning viewpoint, it is evident that additional research efforts are needed, in order to achieve reliable and accurate estimations, especially when multi-level pain assessment is concerned and to enhance clinical adoption of such tools.  In future work, we will focus on the exploration and use of additional biosignals, e.g. EMG or GSR,  in both a unimodal and multimodal setting as well.

\bibliographystyle{splncs04}
\bibliography{library}

\begin{thebibliography}{10}
\providecommand{\url}[1]{\texttt{#1}}
\providecommand{\urlprefix}{URL }
\providecommand{\doi}[1]{https://doi.org/#1}

\bibitem{amirian_kachele_2016}
Amirian, M., Kächele, M., Schwenker, F.: Using radial basis function neural
  networks for continuous and discrete pain estimation from bio-physiological
  signals. vol. 9896 LNAI, pp. 269--284. Springer Verlag (2016).
  \doi{10.1007/978-3-319-46182-3_23}

\bibitem{bartley_fillingim_2013}
Bartley, E.J., Fillingim, R.B.: {Sex differences in pain: a brief review of
  clinical and experimental findings}. British journal of anaesthesia
  \textbf{111}(1),  52--58 (jul 2013). \doi{10.1093/bja/aet127},
  \url{https://pubmed.ncbi.nlm.nih.gov/23794645
  https://www.ncbi.nlm.nih.gov/pmc/articles/PMC3690315/}

\bibitem{williams_craig_2016}
de~C~Williams, A.C., Craig, K.D.: Updating the definition of pain. Pain
  \textbf{157},  2420--2423 (11 2016). \doi{10.1097/j.pain.0000000000000613}

\bibitem{kendall_2018}
Cipolla, R., Gal, Y., Kendall, A.: Multi-task learning using uncertainty to
  weigh losses for scene geometry and semantics. In: 2018 IEEE/CVF Conference
  on Computer Vision and Pattern Recognition. pp. 7482--7491 (2018).
  \doi{10.1109/CVPR.2018.00781}

\bibitem{cordell_2002}
Cordell, W.H., Keene, K.K., Giles, B.K., Jones, J.B., Jones, J.H., Brizendine,
  E.J.: The high prevalence of pain in emergency medical care. American Journal
  of Emergency Medicine  \textbf{20},  165--169 (2002).
  \doi{10.1053/ajem.2002.32643}

\bibitem{dinakar_stillman_2016}
Dinakar, P., Stillman, A.M.: Pathogenesis of pain. Seminars in Pediatric
  Neurology  \textbf{23},  201--208 (8 2016). \doi{10.1016/J.SPEN.2016.10.003}

\bibitem{fariha_ikeura_2020}
Fariha, M.A.Z., Ikeura, R., Hayakawa, S., Tsutsumi, S.: Analysis of
  pan-tompkins algorithm performance with noisy \{ECG\} signals. Journal of
  Physics: Conference Series  \textbf{1532},  12022 (2020).
  \doi{10.1088/1742-6596/1532/1/012022}

\bibitem{gkikas_chatzaki_2022}
Gkikas., S., Chatzaki., C., Pavlidou., E., Verigou., F., Kalkanis., K.,
  Tsiknakis., M.: Automatic pain intensity estimation based on
  electrocardiogram and demographic factors. pp. 155--162. SciTePress (2022).
  \doi{10.5220/0010971700003188}

\bibitem{hadjistavropoulos_craig_2002}
Hadjistavropoulos, T., Craig, K.D.: A theoretical framework for understanding
  self-report and observational measures of pain: A communications model.
  Behaviour Research and Therapy  \textbf{40},  551--570 (5 2002).
  \doi{10.1016/S0005-7967(01)00072-9}

\bibitem{hinduja_canavan_2020}
Hinduja, S., Canavan, S., Kaur, G.: Multimodal fusion of physiological signals
  and facial action units for pain recognition. pp. 577--581. Institute of
  Electrical and Electronics Engineers Inc. (2020).
  \doi{10.1109/FG47880.2020.00060}

\bibitem{huang_feng_2022}
Huang, D., Feng, X., Zhang, H., Yu, Z., Peng, J., Zhao, G., Xia, Z.:
  Spatio-temporal pain estimation network with measuring pseudo heart rate
  gain. IEEE Transactions on Multimedia  \textbf{24},  3300--3313 (2022).
  \doi{10.1109/TMM.2021.3096080}

\bibitem{kachele_thiam_2016}
Kächele, M., Thiam, P., Amirian, M., Schwenker, F., Palm, G.: Methods for
  person-centered continuous pain intensity assessment from bio-physiological
  channels. IEEE Journal on Selected Topics in Signal Processing  \textbf{10},
  854--864 (8 2016). \doi{10.1109/JSTSP.2016.2535962}

\bibitem{martinez_picard_2017}
Lopez-Martinez, D., Picard, R.: Multi-task neural networks for personalized
  pain recognition from physiological signals. vol. 2018-Janua, pp. 181--184.
  Institute of Electrical and Electronics Engineers Inc. (1 2017).
  \doi{10.1109/ACIIW.2017.8272611}

\bibitem{martinez_picard_2018_b}
Lopez-Martinez, D., Picard, R.: Continuous pain intensity estimation from
  autonomic signals with recurrent neural networks. vol.~2018, pp. 5624--5627
  (2018). \doi{10.1109/EMBC.2018.8513575}

\bibitem{martinez_picard_2017_2}
Lopez-Martinez, D., Rudovic, O., Picard, R.: Physiological and behavioral
  profiling for nociceptive pain estimation using personalized multitask
  learning. In: Neural Information Processing Systems (NIPS) Workshop on
  Machine Learning for Health. Long Beach, USA (2017)

\bibitem{pan_tompkins_1985}
Pan, J., Tompkins, W.J.: A real-time qrs detection algorithm. IEEE Transactions
  on Biomedical Engineering  \textbf{BME-32},  230--236 (1985).
  \doi{10.1109/TBME.1985.325532}

\bibitem{Riley_robinson_1998}
3rd Riley, J.L., Robinson, M.E., Wise, E.A., Myers, C.D., Fillingim, R.B.: Sex
  differences in the perception of noxious experimental stimuli: a
  meta-analysis. Pain  \textbf{74},  181--187 (2 1998).
  \doi{10.1016/s0304-3959(97)00199-1}

\bibitem{Rohling_binder_1995}
Rohling, M.L., Binder, L.M., Langhinrichsen-Rohling, J.: Money matters: A
  meta-analytic review of the association between financial compensation and
  the experience and treatment of chronic pain. Health Psychology  \textbf{14},
   537--547 (1995). \doi{10.1037/0278-6133.14.6.537}

\bibitem{khalid_tubbs_2017}
S, K., RS, T.: Neuroanatomy and neuropsychology of pain. Cureus  \textbf{9} (10
  2017). \doi{10.7759/CUREUS.1754}

\bibitem{stewart_2013}
Stewart, G., Panickar, A.: Role of the sympathetic nervous system in pain (12
  2013). \doi{10.1016/j.mpaic.2013.09.003}

\bibitem{subramaniam_dass_2021}
Subramaniam, S.D., Dass, B.: Automated nociceptive pain assessment using
  physiological signals and a hybrid deep learning network. IEEE Sensors
  Journal  \textbf{21},  3335--3343 (2021). \doi{10.1109/JSEN.2020.3023656}

\bibitem{thiam_bellmann_kestler_2019}
Thiam, P., Bellmann, P., Kestler, H.A., Schwenker, F.: Exploring deep
  physiological models for nociceptive pain recognition. Sensors  \textbf{19},
  ~4503 (10 2019). \doi{10.3390/s19204503}

\bibitem{turk_2001}
Turk, D.C., Melzack, R.: The measurement of pain and the assessment of people
  experiencing pain. pp. 3--16 (2011),
  \url{https://psycnet.apa.org/record/2011-03491-001}

\bibitem{biovid_2013}
Walter, S., Gruss, S., Ehleiter, H., Tan, J., Traue, H.C., Crawcour, S.,
  Werner, P., Al-Hamadi, A., Andrade, A.O., Silva, G.M.D.: The biovid heat pain
  database: Data for the advancement and systematic validation of an automated
  pain recognition. pp. 128--131 (2013). \doi{10.1109/CYBConf.2013.6617456}

\bibitem{wang_wei_2020}
Wang, J., Wei, M., Zhang, L., Huang, G., Liang, Z., Li, L., Zhang, Z.: An
  autoencoder-based approach to predict subjective pain perception from
  high-density evoked eeg potentials. vol. 2020-July, pp. 1507--1511. Institute
  of Electrical and Electronics Engineers Inc. (2020).
  \doi{10.1109/EMBC44109.2020.9176644}

\bibitem{werner_hamadi_2014}
Werner, P., Al-Hamadi, A., Niese, R., Walter, S., Gruss, S., Traue, H.C.:
  Automatic pain recognition from video and biomedical signals. pp. 4582--4587.
  Institute of Electrical and Electronics Engineers Inc. (2014).
  \doi{10.1109/ICPR.2014.784}

\bibitem{yu_sun_zhu_2020}
Yu, M., Sun, Y., Zhu, B., Zhu, L., Lin, Y., Tang, X., Guo, Y., Sun, G., Dong,
  M.: Diverse frequency band-based convolutional neural networks for tonic cold
  pain assessment using eeg. Neurocomputing  \textbf{378},  270--282 (2020).
  \doi{10.1016/j.neucom.2019.10.023}

\end{thebibliography}

\end{document}